\documentclass[nonblindrev]{informs3} 

\usepackage{balance}
\usepackage{amsmath}
\usepackage[tight,footnotesize]{subfigure}
\usepackage{graphicx}
\usepackage{bm}
\usepackage{algorithm}
\usepackage[noend]{algorithmic}
\usepackage{multirow}
\usepackage{slashbox}
\usepackage{caption}
\usepackage{amsfonts}

\OneAndAHalfSpacedXII 

\usepackage{natbib}
 \bibpunct[, ]{(}{)}{,}{a}{}{,}%
 \def\bibfont{\small}%
\TheoremsNumberedThrough     

\EquationsNumberedThrough    

\MANUSCRIPTNO{2015} 

\begin{document}


\RUNAUTHOR{Tang et al.}

\RUNTITLE{Lazy Adaptive Submodular Maximization}

\TITLE{Linear-Time Algorithms for Adaptive Submodular Maximization}

\ARTICLEAUTHORS{%
\AUTHOR{Shaojie Tang}
\AFF{Naveen Jindal School of Management, The University of Texas at Dallas}
} 

\ABSTRACT{In this paper, we develop fast algorithms for two stochastic submodular maximization problems. We start with the well-studied adaptive submodular maximization problem subject to a cardinality constraint. We develop the first linear-time algorithm which achieves a $(1-1/e-\epsilon)$ approximation ratio. Notably, the time complexity of our algorithm is $O(n\log\frac{1}{\epsilon})$ (number of function evaluations) which is independent of the cardinality constraint, where $n$ is the size of the ground set. Then we introduce the concept of fully adaptive submodularity, and develop a linear-time algorithm for maximizing a fully adaptive submoudular function subject to a partition matroid constraint. We show that our algorithm achieves a $\frac{1-1/e-\epsilon}{4-2/e-2\epsilon}$ approximation ratio using only $O(n\log\frac{1}{\epsilon})$ number of function evaluations.
}


\maketitle

\section{Introduction}
Submodular maximization is a well-studied topic due to its applications in a wide range of domains including active learning \citep{golovin2011adaptive}, virtual marketing \citep{tang2020influence,yuan2017no}, sensor placement \citep{krause2007near}. Most of existing studies focus on the non-adaptive setting where one must select a group of items all at once subject to some practical constraints such as a cardinality constraint. \cite{nemhauser1978analysis} show that a classic greedy algorithm, which iteratively selects the item that has the largest marginal utility  on top of the previously selected items, achieves $1-1/e$ approximation ratio when maximizing a monotone submodular function subject to a cardinality constraint. However, this algorithm needs $O(n\cdot k)$ function evolutions in order to select all $k$ items, where $n$ is the size of the ground set and $k$ is the cardinality constraint.  Only recently,  \cite{mirzasoleiman2015lazier} improves this time complexity (number of function evaluations) to $O(n\log \frac{1}{\epsilon})$ by proposing a random sampling based stochastic greedy algorithm. Their algorithm can  achieve an approximation ratio  $1-1/e -\epsilon$. Their basic idea can be roughly described as follows: At each round, it draws a small random sample of items, and selects the item, from the random sample, that has the largest marginal utility. They show that if they set the size of the random sample to $\frac{n}{k}\log\frac{1}{\epsilon}$, their algorithm achieves a near-optimal approximation ratio in linear-time. In general, developing fast algorithms for submodular maximization  has received consideration attention during the past several years \citep{leskovec2007cost,badanidiyuru2014fast,mirzasoleiman2016fast,ene2018towards}.

 Recently, \cite{golovin2011adaptive} extends the previous study to the adaptive setting. They propose the problem of adaptive submodular maximization, a natural stochastic variant of the classical non-adaptive submodular maximization problem. They assume that each item is in a particular state drawn from a known prior distribution. The only way to reveal an item's state is to select that item. Note that the decision on selecting an item is selecting an item is irrevocable, that is, we can not discard any item that is previously selected. Their goal is to adaptively select a group of $k$ items so as to maximize the expected utility of an adaptive submodular and adaptive monotone function. There have been numerous research studies on adaptive submodular maximization under different settings \citep{chen2013near,tang2020influence,tang2020price,yuan2017adaptive,fujii2019beyond}. Our first result focuses on adaptive submodular maximization subject to a cardinality constraint. The state-of-the-art for this setting is a simple adaptive greedy policy \citep{golovin2011adaptive} which achieves $1-1/e$ approximation ratio. Similar to the classic non-adaptive greedy algorithm \citep{nemhauser1978analysis}, their adaptive greedy policy needs $O(n\cdot k)$ function evolutions in order to select all $k$ items. One might ask whether is it possible to have a linear-time algorithm for the adaptive setting? In this paper, we answer this question affirmatively and propose the first adaptive policy that achieves $1-1/e-\epsilon$ approximation ratio using only $O(n\log \frac{1}{\epsilon})$ function evaluations. We generalize the non-adaptive stochastic greedy algorithm proposed in \citep{mirzasoleiman2015lazier} to the adaptive setting, and show that  a similar random sampling based technique can be used to evaluate the marginal utility of an item in each round under the adaptive setting. In the second part of this paper, we study a more general optimization problem subject to a partition matroid constraint. To make our problem approachable, we introduce a class of stochastic functions, called \emph{fully adaptive submodular functions}. Then we develop a generalized random sampling based adaptive policy  for maximizing a fully adaptive submoudular function subject to a partition matroid constraint. We show that our algorithm achieves a $\frac{1-1/e-\epsilon}{4-2/e-2\epsilon}$ approximation ratio using only $O(n\log \frac{1}{\epsilon})$ function evaluations.

\emph{Remark:} The other line of work focuses on developing fast parallel algorithms for non-adaptive submodular maximization \citep{balkanski2018adaptive,chekuri2019parallelizing}. They developed approximation algorithms for submodular maximization whose parallel runtime is logarithmic. All algorithms with logarithmic runtime require a good estimate of the optimal value. However, the optimal value is unknown initially, one common approach to resolving this issue is to run multiple instances of algorithms in parallel using different guesses of the optimal value, which ensures that one of those guesses is a good approximation to the optimal value, and then returning the solution with highest value. Unfortunately, our adaptive setting can not afford this type of ``parallelization'' in general. Recall that under the adaptive setting, the decision on selecting an item is irrevocable. When running multiple instances of algorithms under the adaptive setting,  we must keep all solutions returned from those instances. As a result, the existing  studies on parallelization are  not applicable to the adaptive setting.

\section{Preliminaries}
We start by introducing some important notations. In the rest of this paper, we use  $[m]$ to denote the set $\{1, 2, \cdots, m\}$, and we use $|S|$ to denote the cardinality of a set $S$.

\subsection{Items and  States} We are given a set  $E$ of $n$ items, and each item $e\in E$ has a particular state  belonging to $O$.  Let $\phi: E\rightarrow O$ denote a realization of the states of items.   We use $\Phi=\{\Phi_e \mid e\in E\}$ to denote a random realization, where $\Phi_e \in O$ is a random realization of $e$.   The only way to reveal the actual state $\Phi_e$ of an item $e\in E$ is to select that item. We assume there is a known prior probability distribution $p(\phi)=\{\Pr[\Phi=\phi]: \phi\in U\}$ over realizations $U$. A typical adaptive policy works as follows: select the first item and observe its state, then continue to select the next item based on the current observation, and so on. After each selection, we observe some \emph{partial realization} $\psi$ of the states of  some subset of $E$, for example, we are able to observe the partial realization of the states of those items which have been selected. We define the \emph{domain} of $\psi$ as the subset of items involved in $\psi$. For any realization $\phi$ and any  partial realization $\psi$, we say $\psi$ is consistent with $\phi$, denoted $\psi \sim \phi$, if they are equal everywhere in the domain of $\phi$. We say that $\psi$  is a \emph{subrealization} of  $\psi'$, denoted  $\psi \subseteq \psi'$, if $\mathrm{dom}(\psi) \subseteq \mathrm{dom}(\psi)$ and they are equal everywhere in $\mathrm{dom}(\psi)$. Let $p(\phi\mid \psi)$ denote the conditional distribution over realizations conditioned on  a partial realization $\psi$: $p(\phi\mid \psi) =\Pr[\Phi=\phi\mid \psi\sim\Phi ]$. There is a utility function $f$ from a subset of items and their states to a non-negative real number: $f: 2^{E}\times 2^O\rightarrow \mathbb{R}_{\geq0}$.

\subsection{Policies and Problem Formulation} Formally, any adaptive policy  can be represented using a function $\pi$  from a set of partial
realizations to $E$, specifying  which item to select next based on  the current observations.

\begin{definition}[Policy  Concatenation]
Given two policies $\pi$ and $\pi'$,  let $\pi @\pi'$ denote a policy that runs $\pi$ first, and then runs $\pi'$, ignoring the observation obtained from running $\pi$.
\end{definition}

Given any policy $\pi$ and realization $\phi$, let $E(\pi, \phi)$ denote the subset of items selected by $\pi$ under realization $\phi$. The expected  utility $f_{avg}(\pi)$ of a policy $\pi$ is
\begin{equation}
f_{avg}(\pi)=\mathbb{E}_{\Phi\sim p(\phi)}f(E(\pi, \Phi), \Phi)
\end{equation}

Let $\Omega$ denote the set of feasible policies which satisfy some given constraints such as the cardinality constraint. Our goal is to find a feasible policy  $\pi^{opt}$ that maximizes the expected utility:
\[\pi^{opt} \in \arg\max_{\pi \in \Omega} f_{avg}(\pi)\]

\subsection{Adaptive Submodularity, Monotonicity and Fully Adaptive Submodularity }
We start by introducing two notations.
\begin{definition}[Conditional Expected Marginal Utility of an Item]
\label{def:1}
For any partial realization $\psi$ and any item $e$, the conditional expected marginal utility $\Delta(e \mid \psi)$ of $e$ conditioned on $\psi$ is
\[\Delta(e \mid \psi)=\mathbb{E}_{\Phi}[f(\mathrm{dom}(\psi)\cup \{e\}, \Phi)-f(\mathrm{dom}(\psi, \Phi))\mid \Phi \sim \psi]\]
where the expectation is taken over $\Phi$ with respect to $p(\psi\mid \phi)=\Pr(\Phi=\phi \mid \Phi \sim \psi)$.
\end{definition}

\begin{definition}[Conditional Expected Marginal Utility of a Policy]
\label{def:1}
For any partial realization $\psi$ and any item $e$, the conditional expected marginal utility $\Delta(\pi \mid \psi)$ of a policy $\pi$ conditioned on $\psi$ is
\[\Delta(\pi\mid \psi)=\mathbb{E}_{\Phi}[f(\mathrm{dom}(\psi) \cup E(\pi, \Phi), \Phi)-f(\mathrm{dom}(\psi), \Phi)\mid \Phi\sim \psi]\]
where the expectation is taken over $\Phi$ with respect to $p(\psi\mid \phi)=\Pr(\Phi=\phi \mid \Phi \sim \psi)$.
\end{definition}

We next introduce two important concepts proposed in \citep{golovin2011adaptive}, \emph{adaptive submodularity} and \emph{adaptive monotonicity}.
\begin{definition}\citep{golovin2011adaptive}[Adaptive Submodularity]
\label{def:11}
A function $ f: 2^E\times O^E$ is  adaptive submodular with respect to a prior distribution $ p(\phi)$, if for any two partial realizations $\psi$ and $\psi'$ such that $\psi\subseteq \psi'$, the following holds:
\[\Delta(e\mid \psi) \geq \Delta(e\mid \psi') \]
\end{definition}

\begin{definition}\citep{golovin2011adaptive} [Adaptive Monotonicity]
\label{def:2}
A function $ f: 2^E\times O^E$ is  adaptive monotone with respect to a prior distribution $ p(\phi)$, if for any realization $\psi$, the following holds:
\[\Delta(e\mid \psi) \geq 0 \]
\end{definition}

At last, we introduce the concept of \emph{fully adaptive submodularity}, a stricter condition than the adaptive submodularity.

\begin{definition}[Fully Adaptive Submodularity]
\label{def:1}
For any subset of items $V\subseteq E$ and any integer $a\in[|V|]$, let  $\Omega(V,a)$  denote the set of policies which are allowed to select at most $a$ items only from $V$. A function $f$ is fully adaptive submodular with respect to a prior distribution $ p(\phi)$, if for any two partial realizations $\psi$ and $\psi'$ such that $\psi\subseteq \psi'$, and any subset of items $V\subseteq E$ and any $a\in [|V|]$, the following holds:
\[\max_{\pi\in \Omega(V,a)} \Delta(\pi \mid \psi) \geq \max_{\pi\in \Omega(V,a)} \Delta(\pi \mid \psi') \]
\end{definition}

In the above definition, $V$  can be any single item, thus any fully adaptive submodular function must be adaptive submodular according to Definition \ref{def:11}.

\section{Adaptive Stochastic Greedy Policy}
We first study the adaptive submodular maximization problem subject to a cardinality constraint. Under this setting, we say a policy is feasible if it selects at most $k$ items under any realization. Formally, let $\Omega=\{\pi\mid  \forall \phi, |E(\pi, \phi)|\leq k\}$ denote the set of all feasible policies subject to a cardinality constraint $k$, our goal is to find a policy from $\Omega$ that maximizes the expected utility.

We next present our adaptive policy \emph{Adaptive Stochastic Greedy}, denoted by  $\pi^{asg}$. The details of our algorithm are listed in Algorithm \ref{alg:LPP1}. Similar to the classic adaptive greedy algorithm, $\pi^{asg}$ runs round by round: Starting with an empty set and  at each round, it selects one item that maximizes the expected marginal utility based on the current observation. What is different from the adaptive greedy algorithm, however, is that at each round $r\in[k]$, $\pi^{asg}$ first samples a set $S_r$ of size $\frac{n}{k}\log\frac{1}{\epsilon}$ uniformly at random and then selects the item with the largest conditional expected marginal utility from $S_r$. Our approach is a natural extension of the \emph{Stochastic Greedy} algorithm \citep{mirzasoleiman2015lazier}, the first linear-time algorithm for maximizing a monotone submodular function under the non-adaptive setting, we generalize their results to the adaptive setting. We will show that $\pi^{asg}$  achieves $(1-1/e-\epsilon)$ approximation ratio for maximizing an adaptive monotone and submodular function, and it has linear running time independent of the cardinality constraint $k$. It was worth noting that the technique of lazy updates \citep{minoux1978accelerated}  can be used to further accelerate the computation of our algorithms in practice.

\begin{algorithm}[hptb]
\caption{Adaptive Stochastic Greedy $\pi^{asg}$}
\label{alg:LPP1}
\begin{algorithmic}[1]
\STATE $A=\emptyset; r=1$.
\WHILE {$r \leq k$}
\STATE observe $\psi_r$;
\STATE $S_r\leftarrow$ a random set sampled uniformly at random  from $E$;
\STATE $e_r\leftarrow \arg\max_{e \in S_r}\Delta(e\mid \psi_r)$;
\STATE $A\leftarrow A\cup \{e_r\}$; $r\leftarrow r+1$;
\ENDWHILE
\RETURN $A$
\end{algorithmic}
\end{algorithm}

We next present the main theorem.
\begin{theorem}
\label{thm:1}
If $f$ is adaptive submodular and adaptive monotone, then the Adaptive Stochastic Greedy policy $\pi^{asg}$ achieves a $(1-1/e-\epsilon)$ approximation ratio in expectation  with $O(n \log  \frac{1}{\epsilon})$ function evaluations.
\end{theorem}

\emph{Proof:} To prove this theorem, we follow an argument similar to the one from \citep{mirzasoleiman2015lazier} for the proof of Theorem 1, but extended to the adaptive setting. We first prove the time complexity of $\pi^{asg}$. Recall that we set the size of $S_r$ to $\frac{n}{k}\log\frac{1}{\epsilon}$, thus the total number of function evaluations is at most $k\times \frac{n}{k}\log\frac{1}{\epsilon}=n \log  \frac{1}{\epsilon}$.

 Before proving the approximation ratio of $\pi^{asg}$, we first provide a preparation lemma as follows.
\begin{lemma}
\label{lem:a}
Given any partial realization $\psi$, let $A^*\in \arg\max_{A\subseteq  E, |A|= k} \sum_{e\in A} \Delta(e\mid \psi)$ denote the $k$ largest items in terms of the expected marginal utility conditioned on having observed $\psi$. Assume $R$ is sampled  uniformly at random  from $E$, and the size of $R$ is $|R|=\frac{n}{k}\log\frac{1}{\epsilon}$. We have $\Pr[R\cap  A^*\neq \emptyset]\geq 1-\epsilon$.
\end{lemma}

The proof of Lemma \ref{lem:a} is moved to Appendix. Given Lemma \ref{lem:a} in hand, now we are ready to bound the approximation ratio of $\pi^{asg}$. Let $A_r=\{e_1, e_2, \cdots, e_r\}$ denote the first $r$ items selected by  $\pi^{asg}$, and $\psi_r$ denote the partial realization observed till round $r$, e.g., $\mathrm{dom}(\psi_r)=A_r$. Our goal is to estimate the increased utility  $f(\pi^{asg}_{r+1})-f(\pi^{asg}_r)$ during one round of our algorithm where $\pi^{asg}_r$ denotes a policy that runs $\pi^{asg}$ until it selects $r$ items.
Given a partial realization $\psi_r$ at round $r$, denote by $A_r^*\in \arg\max_{A\subseteq  E, |A|= k} \sum_{e\in A} \Delta(e\mid \psi_r)$ the $k$ largest items in terms of the expected marginal utility conditioned on having observed $\psi_r$. Recall that  $S_r$ is sampled uniformly at random from $E$, each item in $A_r^*$ is equally likely to be contained in $S_r$. Moreover, the size of $S_r$ is set to $\frac{n}{k}\log\frac{1}{\epsilon}$. It follows that
\begin{eqnarray*}
\mathbb{E}_{e_r}[\Delta(e_r\mid \psi_r)]&=&\frac{1}{k}\Pr[S_r\cap  A_r^* \neq \emptyset] \sum_{e\in A_r^*} \Delta(e\mid \psi_r)\\
&\geq&\frac{1}{k}(1-\epsilon)\sum_{e\in A_r^*} \Delta(e\mid \psi_r)
\end{eqnarray*} where the inequality is due to Lemma \ref{lem:a}. Then we have
\[\mathbb{E}_{e_r}[\Delta(e_r\mid \psi_r)] \geq (1-\epsilon) \frac{1}{k}\sum_{e\in A^*_r} \Delta(e\mid \psi_r)\geq (1-\epsilon) \frac{1}{k} \Delta(\pi^{opt}\mid \psi_r)\] where the second inequality is due to Lemma 6 in \citep{golovin2011adaptive}.
Therefore
\[\mathbb{E}_{e_r}[f(\pi^{asg}_{r+1})-f(\pi^{asg}_r)\mid \psi_r] = \mathbb{E}_{e_r}[\Delta(e_r\mid \psi_r)] \geq (1-\epsilon) \frac{1}{k} \Delta(\pi^{opt}\mid \psi_r)\]

Then we have
\begin{eqnarray}
f(\pi^{asg}_{r+1})-f(\pi^{asg}_r)=\mathbb{E}_{e_r, \psi_r}[\Delta(e_r\mid \psi_r)]  &\geq& (1-\epsilon) \frac{1}{k} \mathbb{E}_{ \psi_r}[ \Delta(\pi^{opt}\mid \psi_r)]\\
&=& (1-\epsilon) \frac{1}{k} ( f(\pi^{asg}_r@\pi^{opt})-f(\pi^{asg}_r))\\
&\geq&(1-\epsilon)\frac{1}{k} (f(\pi^{opt})-f(\pi^{asg}_r))\label{eq:c}
\end{eqnarray}
Note that the first expectation is taken over two sources of randomness: one source is the randomness in choosing $e_r$, and the other source is the randomness in the partial realization $\psi_r$. The second inequality is due to $f$ is adaptive monotone.  Based on (\ref{eq:c}), we have $f(\pi^{asg})=f(\pi^{asg}_k)\geq (1-1/e-\epsilon)f(\pi^{opt})$ using induction. $\Box$

\section{Generalized Adaptive Stochastic Greedy Policy}
We next study the fully adaptive submodular maximization problem subject to a partition matroid constraint. Let $B_1, B_2, \ldots, B_b$ be a collection of disjoint subsets
of $E$. Given a set of $b$ integers $\{d_i\mid i\in [b]\}$, define  $\Omega=\{\pi\mid  \forall \phi, \forall i\in[b], |E(\pi, \phi) \cap B_i|\leq d_i\}$ as the set of all feasible policies. Our goal is to find a feasible policy that maximizes an adaptive monotone and fully adaptive submodular function.

\subsection{Locally Greedy Policy}
 Before presenting our linear-time algorithm, we first introduce a \emph{Locally Greedy} policy $\pi^{local}$. The basic idea of $\pi^{local}$ is as follows: Starting with an empty set, $\pi^{local}$ first selects $d_1$ number of items from $B_1$ in a greedy manner, i.e., iteratively adds items that maximize the conditional expected marginal utility conditioned on the realizations of  already selected items; $\pi^{local}$ then selects $d_2$ number of items from $B_2$ in the same greedy manner, and so on. Note that  $\pi^{local}$ does not rely on any specific ordering of $B_i$, and this motivates the term ``locally greedy'' that we use to name this policy. We can view $\pi^{local}$ as an adaptive version of the locally greedy algorithm proposed in  \citep{fisher1978analysis}.

 We first introduce some additional concepts.  Given a policy $\pi$, for any $i\in [b]$ and $j\in [d_i]$, we define 1) its \emph{level-$(i,j)$-truncation} $\pi_{ij}$ as a policy that runs $\pi$ until it selects  $j$ items from $B_i$ and 2) its \emph{strict level-$(i,j)$-truncation} $\pi_{\leftarrow ij}$ as a policy that runs $\pi$ until right before it selects  the $j$-th item  from $B_i$. It is clear that $f_{avg}(\pi)=\sum_{i\in[b], j\in[d_i]}(f_{avg}(\pi_{ij})-f_{avg}(\pi_{\leftarrow ij}))$.

We next show that if $f$ is adaptive submodular and adaptive monotone, the expected utility of $\pi^{local}$ is at least half of the optimal solution.
\begin{lemma}
\label{lem:6}
If $f$ is adaptive submodular and adaptive monotone, then  $f_{avg}(\pi^{local}) \geq  f_{avg}(\pi^{opt})/2$.
\end{lemma}

\emph{Proof:} Consider a policy $\pi^{local} @\pi^{opt}_{\leftarrow ij}$ that runs $\psi^{local}$ first, then runs the strict level-$(i,j)$-truncation of $\pi^{opt}$. Let $\psi^{l@o_{\leftarrow ij}}$ denote the partial realization obtained after running $\pi^{local} @\pi^{opt}_{\leftarrow ij}$. Based on this notation, we use $\psi^{l@o_{\leftarrow ij}}_{b_{\leftarrow ij }} \subseteq \psi^{l@o_{\leftarrow ij}}$ to denote a subrealization of $\psi^{l@o_{\leftarrow ij}}$ obtained after running the strict level-$(i,j)$-truncation of $\pi^{local}$. Conditioned on $\psi^{l@o_{\leftarrow ij}}$, assume $\pi^{local}$ selects $e^{local} _{ij}$ as the $j$-th item from $B_i$, and $\pi^{opt}$ selects $e^{opt}_{ij}$ as the $j$-th item from $B_i$. We first bound the expected marginal utility $\Delta(e^{local} _{ij}\mid \psi^{l@o_{\leftarrow ij}}_{b_{\leftarrow ij }})$ of $e^{local} _{ij}$ conditioned on having observed $\psi^{l@o_{\leftarrow ij}}_{b_{\leftarrow ij }}$.
\begin{eqnarray}
\Delta(e^{local} _{ij}\mid \psi^{l@o_{\leftarrow ij}}_{b_{\leftarrow ij }})&=& \max_{e\in B_i}\Delta(e \mid \psi^{l@o_{\leftarrow ij}}_{b_{\leftarrow ij }})\\
&\geq& \max_{e\in B_i}\Delta(e \mid \psi^{l@o_{\leftarrow ij}})\\
&\geq& \Delta(e^{opt}_{ij} \mid \psi^{l@o_{\leftarrow ij}})\\
&=& f_{avg}(\pi^{local} @\pi^{opt}_{ij}\mid \psi^{l@o_{\leftarrow ij}})-f_{avg}(\pi^{local} @\pi^{opt}_{\leftarrow ij}\mid \psi^{l@o_{\leftarrow ij}}) \label{eq:d}
\end{eqnarray}
The first equality is due to $\pi^{local}$ selects an item that maximizes the conditional expected marginal utility. The first inequality is due to the assumption that $f$ is adaptive submodular.

Taking the expectation of $\Delta(e^{local} _{ij}\mid \psi^{l@o_{\leftarrow ij}}_{b_{\leftarrow ij }})$ over $\psi^{l@o_{\leftarrow ij}}$, we have
\begin{eqnarray} &&\mathbb{E}_{\psi^{l@o_{\leftarrow ij}}}[\Delta(e^{local} _{ij}\mid \psi^{l@o_{\leftarrow ij}}_{b_{\leftarrow ij }})]
= f_{avg}(\pi^{local}_{ij})-f_{avg}(\pi^{local}_{\leftarrow ij})\\
&\geq& \mathbb{E}_{\psi^{l@o_{\leftarrow ij}}}[f_{avg}(\pi^{local} @\pi^{opt}_{ij}\mid \psi^{l@o_{\leftarrow ij}})-f_{avg}(\pi^{local} @\pi^{opt}_{\leftarrow ij}\mid \psi^{l@o_{\leftarrow ij}})]\label{eq:f}\\
&=& f_{avg}(\pi^{local} @\pi^{opt}_{ij})-f_{avg}(\pi^{local} @\pi^{opt}_{\leftarrow ij}) \label{eq:1}
\end{eqnarray}
where (\ref{eq:f}) is due to (\ref{eq:d}). Because for any policy $\pi$, we have $f_{avg}(\pi)=\sum_{i\in[b], j\in[d_i]}(f_{avg}(\pi_{ij})-f_{avg}(\pi_{\leftarrow ij}))$, it follows that
\begin{eqnarray}
f_{avg}(\pi^{local})&=&\sum_{i\in[b], j\in[d_i]}(f_{avg}(\pi^{local}_{ij})-f_{avg}(\pi^{local}_{\leftarrow ij}))\\
&\geq& \sum_{i\in[b], j\in[d_i]} (f_{avg}(\pi^{local} @\pi^{opt}_{ij})-f_{avg}(\pi^{local} @\pi^{opt}_{\leftarrow ij}))\\
&=& f_{avg}(\pi^{local} @\pi^{opt}) - f_{avg}(\pi^{local})\\
&\geq&  f_{avg}(\pi^{opt}) - f_{avg}(\pi^{local})
\end{eqnarray}
The first inequality is due to (\ref{eq:1}), and the second inequality is due to the assumption that $f$ is adaptive monotone.
Then we have $ f_{avg}(\pi^{local}) \geq  f_{avg}(\pi^{opt})/2$. $\Box$

\emph{Remark:} We believe that  Lemma \ref{lem:6}  is of independent interest in the field of adaptive submodular maximization under matroid constraints. In particular,  \cite{golovin2011adaptive1} shows that a classic adaptive greedy algorithm, which iteratively selects an item with the largest contribution to the previously selected items, achieves an approximation ratio $1/2$. This approximation factor is nearly optimal. When applied to our problem, their algorithm needs $O((\sum_{i\in[b]} d_i)n)$ function evaluations. Since our locally greedy policy $\pi^{local}$  does not rely on any specific ordering of $B_i$, it requires $O(\sum_{i\in[b]} d_i |B_i|)$ function evaluations.

\subsection{Generalized Adaptive Stochastic Greedy Policy}
Although $\pi^{local}$ requires less function evaluations than the classic greedy algorithm, its runtime is still dependent on $d_i$, the cardinality constraint for each group $B_i$. We next present a linear-time adaptive policy \emph{Generalized Adaptive Stochastic Greedy}, denoted $\pi^{gasg}$, whose runtime is independent of $d_i$. The basic idea of $\pi^{gasg}$ is similar to $\pi^{local}$, except that we leverage the adaptive stochastic greedy policy $\pi^{asg}$ to select a group of $d_i$ items from $B_i$ for each $i\in[b]$. A detailed description of $\pi^{gasg}$ can be found in Algorithm \ref{alg:LPP2}.  We next brief the idea of $\pi^{gasg}$:   Starting with an empty set, $\pi^{gasg}$ first selects $d_1$ number of items from $B_1$ using $\pi^{asg}$, i.e., first sampling a random set $S_{11}$ with size $\frac{|B_1|}{d_1}\log\frac{1}{\epsilon}$ uniformly at random from $B_1$, then selects an item  with the largest conditional expected marginal utility from $S_{11}$, selecting the next item from a newly sampled set  $S_{12}$ in the same greedy manner, and so on; $\pi^{local}$ then selects $d_2$ number of items from $B_2$ using $\pi^{asg}$, where the size of the random set is set to $\frac{|B_2|}{d_2}\log\frac{1}{\epsilon}$,  conditioned on the current observation, and so on. Similar to   $\pi^{local}$,  $\pi^{gasg}$ does not rely on any specific ordering of $B_i$.

\begin{algorithm}[hptb]
\caption{Generalized Adaptive Stochastic Greedy $\pi^{gasg}$}
\label{alg:LPP2}
\begin{algorithmic}[1]
\STATE $A=\emptyset; i=1; j=1$.
\WHILE {$i \leq b$}
\WHILE {$j \leq d_i$}
\STATE observe the current partial realization $\psi^{gasg}_{\leftarrow ij}$;
\STATE $S_{ij}\leftarrow$ a random set, with size $\frac{|B_i|}{d_i}\log\frac{1}{\epsilon}$, sampled  uniformly at random from $B_i$;
\STATE $e^{gasg}_{ij}\leftarrow \arg\max_{e \in S_{ij}}\Delta(e\mid \psi^{gasg}_{\leftarrow ij})$;
\STATE $A\leftarrow A\cup \{e^{gasg}_{ij}\}$; $j\leftarrow j+1$;
\ENDWHILE
$j\leftarrow 1$; $i\leftarrow i+1$;
\ENDWHILE
\RETURN $A$
\end{algorithmic}
\end{algorithm}

We next analyze the performance bound of $\pi^{gasg}$. We start by showing that the expected utility of $\pi^{gasg}$ is at least $\frac{1-1/e-\epsilon}{2-1/e-\epsilon}$ times the expected utility of $\pi^{local}$.
\begin{lemma}
\label{lem:5}
If $f$ is full adaptive submodular and adaptive monotone, then  $f_{avg}(\pi^{gasg}) \geq \frac{1-1/e-\epsilon}{2-1/e-\epsilon} f_{avg}(\pi^{local})$.
\end{lemma}
\emph{Proof:} For ease of presentation, we use $\pi^{local}_i$ (resp. $\pi^{gasg}_i$) to denote a policy that runs $\pi^{local}$ (resp. $\pi^{gasg}$) until it selects all $\sum_{z\in[i]}d_z$ items from $B_1, B_2, \ldots, B_i$. Consider a policy $\pi^{gasg} @\pi^{local}_i$ that runs $\psi^{gasg}$ first, then runs  $\pi^{local}_i$. Let $\psi^{g@l_i}$ denote the partial realization obtained after running $\pi^{gasg} @\pi^{local}_i$. We use $\psi^{g@l_i}_{g_i}\subseteq \psi^{g@l_i}$ to denote a subrealization of $\psi^{g@l_i}$ obtained after running $\pi^{gasg}_i$.   For a fixed $i\in[b]$, let $\Omega(B_i, d_i)$ denote a set of polices which are allowed to select at most $d_i$ items only from $B_i$. For a given $\psi^{g@l_i}$, we first bound the conditional expected marginal utility $\Delta(\pi^{gasg}_{i+1}\mid \psi^{g@l_i}_{g_i})$ of the $d_{i+1}$ items  selected  from $B_{i+1}$ by $\pi^{gasg}$. Let $\alpha=1-1/e-\epsilon$,
\begin{eqnarray}
\Delta(\pi^{gasg}_{i+1}\mid \psi^{g@l_i}_{g_i})&\geq&\alpha\max_{\pi\in \Omega(B_{i+1}, d_{i+1})}\Delta(\pi \mid \psi^{g@l_i}_{g_i})\\
&\geq& \alpha\max_{\pi\in \Omega(B_{i+1}, d_{i+1})} \Delta(\pi \mid \psi^{g@l_i})\\
&\geq& \alpha \Delta(\pi^{gasg} @ \pi^{local}_{i+1} \mid \psi^{g@l_i})\\
&=& \alpha (f_{avg}(\pi^{gasg} @\pi^{local}_{i+1}\mid  \psi^{g@l_i})-f_{avg}(\pi^{gasg} @\pi^{local}_{i}\mid  \psi^{g@l_i})) \label{eq:2}
\end{eqnarray}
The first inequality is due to Theorem \ref{thm:1} and the fact that we use $\pi^{asg}$ to select $d_i$ items from each group $B_i$. The second inequality is due to $f$ is adaptive monotone.

Taking the expectation of $\Delta(\pi^{gasg}_{i+1}\mid \psi^{g@l_i}_{g_i})$ over $\psi^{g@l_i}$,  we have
\begin{eqnarray}
\mathbb{E}_{\psi^{g@l_i}}[\Delta(\pi^{gasg}_{i+1}\mid \psi^{g@l_i}_{g_i})]&=& f_{avg}(\pi^{gasg}_{i+1})-f_{avg}(\pi^{gasg}_{i}) \\
&\geq& \mathbb{E}_{\psi^{g@l_i}}[ \alpha (f_{avg}(\pi^{gasg} @\pi^{local}_{i+1}\mid  \psi^{g@l_i})-f_{avg}(\pi^{gasg} @\pi^{local}_{i}\mid  \psi^{g@l_i}))]\\
&=& \alpha (f_{avg}(\pi^{gasg} @\pi^{local}_{i+1})-f_{avg}(\pi^{gasg} @\pi^{local}_{i}))\label{eq:3}
\end{eqnarray}
The inequality is due to (\ref{eq:2}).

Because $f_{avg}(\pi^{gasg})=\sum_{i\in[b-1]} (f_{avg}(\pi^{gasg}_{i+1})-f_{avg}(\pi^{gasg}_{i}))$ and $f_{avg}(\pi^{gasg} @\pi^{local})=f_{avg}(\pi^{gasg}) + \sum_{i\in[b-1]}f_{avg}(\pi^{gasg} @\pi^{local}_{i+1})-f_{avg}(\pi^{gasg} @\pi^{local}_{i})$, we have
\begin{eqnarray}
f_{avg}(\pi^{gasg})&=&\sum_{i\in[b-1]} (f_{avg}(\pi^{gasg}_{i+1})-f_{avg}(\pi^{gasg}_{i})) \\
&\geq& \alpha\sum_{i\in[b-1]} (f_{avg}(\pi^{gasg} @\pi^{local}_{i+1})-f_{avg}(\pi^{gasg} @\pi^{local}_{i}))\\
&=&\alpha ( f_{avg}(\pi^{gasg} @\pi^{local})-f_{avg}(\pi^{gasg}))\\
&\geq& \alpha ( f_{avg}(\pi^{local})-f_{avg}(\pi^{gasg}))
\end{eqnarray}
The first inequality is due to (\ref{eq:3}) and the second inequality is due to $f$ is adaptive monotone. It follows that $f_{avg}(\pi^{gasg}) \geq \frac{\alpha}{1+\alpha}f_{avg}(\pi^{local})=\frac{1-1/e-\epsilon}{2-1/e-\epsilon}f_{avg}(\pi^{local})$. $\Box$

We next present the main theorem and show that the approximation ratio of $\pi^{gasg}$ is at least $ \frac{1-1/e-\epsilon}{4-2/e-2\epsilon}$ and its running time is $O(n\log\frac{1}{\epsilon})$ (number of function evaluations).
\begin{theorem}
If $f$ is fully adaptive submodular and adaptive monotone, then  $f_{avg}(\pi^{gasg}) \geq \frac{1-1/e-\epsilon}{4-2/e-2\epsilon} f_{avg}(\pi^{opt})$, and $\pi^{gasg}$ uses at most $O(n\log\frac{1}{\epsilon})$ function evaluations.
\end{theorem}

\emph{Proof:} Lemma \ref{lem:6} and Lemma \ref{lem:5} together imply that $f_{avg}(\pi^{gasg}) \geq \frac{1-1/e-\epsilon}{4-2/e-2\epsilon} f_{avg}(\pi^{opt})$. Recall that in Algorithm \ref{alg:LPP2}, we set the size of the random set  $S_{ij}$ to $\frac{|B_i|}{d_i}\log\frac{1}{\epsilon}$, thus the total number of function evaluations is $\sum_{i\in[b], j\in{[d_i]}}\frac{|B_i|}{d_i}\log\frac{1}{\epsilon}= n\log\frac{1}{\epsilon}$. $\Box$

\section{Conclusion}
In this paper, we develop the first linear-time algorithms for the adaptive submodular maximization problem subject to a cardinality constraint. We also study the fully adaptive submodular maximization problem subject to a partition matroid constraint, and develops a linear-time algorithm whose approximation ratio is a constant. In the future, we would like to develop fast algorithms subject to more general constraints such as knapsack constraint and general matroid constraints.

\section*{Appendix}
Proof of Lemma \ref{lem:a}:   We first provide a lower bound the probability that $R\cap  A^* =\emptyset$.
\begin{eqnarray}
\Pr[R\cap  A^* = \emptyset]&=&{(1-\frac{|A^*|}{|E|})}^{|R|} ={(1-\frac{k}{n})}^{|R|}\leq e^{-|R|\frac{k}{n}}
\end{eqnarray}

It follows that $\Pr[R\cap  A^* \neq \emptyset]\geq 1- e^{-\frac{|R|}{n}k}$. Because we assume $|R|=\frac{n}{k}\log\frac{1}{\epsilon}$, we have \begin{equation}\label{eq:a}\Pr[R\cap  A^*\neq \emptyset]\geq 1-e^{-|R|\frac{k}{n}} \geq 1-  e^{-\frac{\frac{n}{k}\log\frac{1}{\epsilon}}{n}k}\geq 1-\epsilon
\end{equation}

\bibliographystyle{ijocv081}
\bibliography{reference}




\end{document}